\documentclass[runningheads,anonymous]{llncs}
\usepackage{graphicx}
\usepackage{multirow}
\usepackage{chngpage}
\usepackage{amsmath}

\begin{document}
\title{Learning Behavioral Representations from Wearable Sensors}%
%
\author{Nazgol Tavabi\inst{1} \and
Homa Hosseinmardi\inst{1} \and
Jennifer L. Villatte\inst{2} \and
Andr\'es Abeliuk\inst{1} \and
Shrikanth Narayanan\inst{1} \and
Emilio Ferrara\inst{1} \and
Kristina Lerman\inst{1}}
%
%
\institute{USC Information Sciences Institute, University of Washington}
%
\maketitle              
\begin{abstract}
Continuous collection of physiological data from wearable sensors enables temporal characterization of individual behaviors. 
Understanding the relation between an individual's behavioral patterns and psychological states can help identify strategies to improve quality of life. 
One challenge in analyzing physiological data is extracting the underlying behavioral states from the temporal sensor signals and interpreting them. Here, we use a non-parametric Bayesian approach to model sensor data from multiple people and discover the dynamic behaviors they share. We apply this method to data collected from sensors worn by a population of hospital workers and show that the learned states 
can cluster participants into meaningful groups and better predict their cognitive and psychological states. 
This method offers a way to learn interpretable compact behavioral representations from multivariate sensor signals. 

\end{abstract}
\section{Introduction}
\begin{figure*}[!t]
\centering
\includegraphics[width=\linewidth]{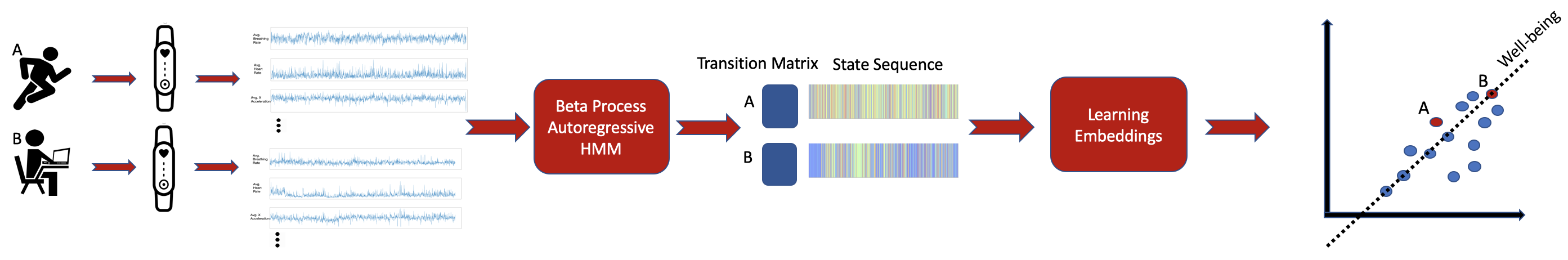}\caption{Overview of the modeling framework. Sensor data collected from participants A and B is fed into BP-AR-HMM model which outputs an HMM per participant, where states are shared among participants. Output from BP-AR-HMM model is used to learn embeddings, which are later used to predict personal attributes.}
\label{fig:framework}
\end{figure*}

Advances in sensing technologies have made wearable sensors more accurate and widely available, enabling continuous and unobtrusive acquisition of physiological data
. This data potentially allows quantitative characterization of human behavior, which provides a basis to 
assess an individual's health~\cite{aral2017exercise} and psychological well-being~\cite{wang2014studentlife}. 
To make sense of physiological data, a  number of challenges have to be met. Sensor data is dynamic 
and covers different time periods. Another challenge is the heterogeneity of data collected from a population of differently-behaving individuals. 
These issues make aggregating and modeling sensor data  challenging.  
Researchers have used Hidden Markov Models (HMMs) to address some of these challenges and effectively capture temporal trends within physiological data~\cite{novak2004morphology,pierson2018modeling}
. HMMs are a family of generative probabilistic models for sequential data, which can be represented as a Markov process with 
latent, or ``hidden," states. 
These hidden states 
capture common patterns that determine the dynamics of the data. One weakness of traditional HMMs is that they constrain  the model to a predefined number of states. When learning dynamic behaviors from  multiple physiological signals
, it may be difficult to enumerate the states best representing the data without making strong assumptions. 
To address this challenge, we propose to use a non-parametric Markov Switching Autoregressive model~\cite{fox2014joint}---the \textit{Beta Process Autoregressive HMM}. 
The number of states in this model 
is learned from the data: if an unusual new pattern appears, 
another state is added to model that segment. This is beneficial in cases where there is a malfunction or noise in the sensors. By assigning a separate state to that segment, we can identify and disregard that state.  
We apply the model to physiological data collected from about 200 workers at a hospital 
and show that the proposed model learns states that correspond to shared behaviors of the workers. 
We use these behavioral representations to better understand and analyze the data, group similar individuals together, 
and as features to predict their personality traits. 
Figure \ref{fig:framework} shows an overview of our framework.

\section{Related Work}
Learning compact representations becomes especially important when dealing with physiological data. In these types of problems we usually don’t have a large number of participants; however, for each participant we have rich longitudinal data. Since we have one label per data stream--participant--our training data is limited to the number of participants in the study, in this case 180 data points. Complex models, such as those learned by deep neural networks, tend to overfit on small training data, and thus are not applicable for these problems. 

Hidden Markov Models (HMMs) represent temporal trends and dynamics of time series using states and transition probabilities. 
In prior research, 
HMM models are learned on each time series independently. 
Consequently, the learned states cannot be compared across  different representations.  
In addition to HMM models,  other recent methods for temporal modeling  suffer from the same shortcoming
; for example, the approach by Hallac et al.\cite{hallac2017toeplitz}, which offers a new perspective on clustering subsequences of multivariate time series, cannot be used in learning representations across multiple signals. 
Another shortcoming of standard HMM models is that the number of states must be fixed a priori. Recent Bayesian approaches overcome these constraints by allowing infinitely many potential states using Beta process, which are shared among all time series\cite{fox2014joint}. 
This approach has successfully been applied to different applications\cite{houpt2018unsupervised,tavabi2019characterizing}. 
Another line of work allows infinite potential states in a Hidden Markov Model \cite{beal2002infinite} by using the Dirichlet process as prior over the hidden states, but the model is again designed to capture each time series independently. 
Tensor decomposition is another approach in finding shared latent features among multiple time series.  Tensor-based methods have been used in different fields 
including behavioral modeling~\cite{hosseinmardi2018discovering,hosseinmardi2018tensor}. 
A popular tensor decomposition method is
Parafac2~\cite{harshman1970foundations}, 
which offers multilinear higher-order decomposition that can handle missing values and different length time series. Parafac2 itself is considered a traditional method, however there are multiple works published in recent years on improving its inference, imposing constraints, etc \cite{jorgensen2018probabilistic,cohen2018nonnegative}. 
Similarly, a recent work \cite{wu2018random} proposes Random Warping Series (RWS), a model based on Dynamic Time Warping, to 
embed different length, multivariate time series in a multi-dimensional space. 

\section{Methods}
\label{sec:methods}



One of the most popular tools for studying multivariate time series is vector autoregressive (VAR) model~\cite{monbet2017sparse}. In a VAR model of lag $r$, each variable is a linear function of itself and the other variable's $r$ previous values.  
However, such models cannot describe time series with changing behaviors. 
In order to model such cases, Markov switching autoregressive models, which are a generalization of autoregressive and Hidden Markov Models, are used. 
In this paper, we use a generative model  proposed by \cite{fox2014joint}, called \textit{Beta Process Autoregressive HMM} (BP-AR-HMM), to discover states or behaviors shared by different time series. Based on the proposed model, the entire set of time series can be described by the globally-shared states, or behaviors, where each time series is associated with a subset of the states. Behaviors associated with different time series can be represented by a binary matrix $F$, where $F_{ij} = 1$ means time series $i$ is associated with behavior $j$.  Given matrix $F$, each time series is modeled as a separate hidden Markov model with the states that it exhibits.
Also, each state is modeled using a different autoregressive process.  
Since the number of such states in the data is not known in advance, the Beta process is used~\cite{thibaux2007hierarchical}. A Beta process allows infinite number of behaviors but encourages sparse representations. 
This process is also known as the \textit{Indian Buffet Process} 
which can be best understood with the following metaphor involving a sequence of customers (time series) selecting dishes (features) from an infinitely large buffet. The $n$-th customer  selects dish $k$ with probability $m_k/n$, where $m_k$ is the popularity of the dish. 
S/He then selects $Poisson(\alpha/n)$ new dishes. With this approach, 
the feature space increases if the data cannot be faithfully represented with the already defined states. However, the probability of adding new states decreases according to $Poisson(\alpha/n)$.  
For posterior computations the original work is referenced \cite{fox2014joint}.

\subsection{Measuring Distance} 
\label{sec:clustering}
When applied to physiological signals, the  
generative model described above learns a hidden Markov model for each participant. 
In this section we propose two different methods for measuring the distance between HMMs, participants. 

 \textbf{Likelihood Distance: } To define a similarity measure between two HMMs, one could measure the probability of their state sequences 
having been generated by the same process. Since  each signal is associated with  its distinct generative process, we measure  state sequences' 
similarity as the likelihood that sequence  ($S_\lambda$) was generated by $\lambda^\prime$, the process that gave rise to ($S_{\lambda^\prime}$) denoted by \textbf{  $p_{\lambda^\prime}(S_\lambda)$} ,  and the likelihood that $S_{\lambda^\prime}$ was generated by $\lambda$.  
We average the two likelihoods to symmetrize the similarity measure.
The likelihood $p_{\lambda^\prime}(S_\lambda)$ is computed using the learned transition matrix of $\lambda^\prime$ and Markov process assumption $ z_t|z_{t-1} \sim\ \pi_{z_{t-1},\lambda^\prime}$
, where $\pi_{z_{t-1},\lambda^\prime}$ is a row of the transition matrix corresponding to state $z_{t-1}$. Because a longer time series would automatically have a smaller likelihood with this approach, we normalize them by dividing $p_{\lambda^\prime}(S_\lambda)$ to ${\frac{1}{K}}^L$, $K$ being the number of states and $L$ being the length of the time series. 

\textbf{Viterbi Distance: } Distance between different HMMs could be also computed with Viterbi distance proposed in \cite{falkhausen1995calculation}, 
which is estimated as follows: 

\begin{equation}
d_{Vit}^\sim(\lambda, \lambda^\prime) = \sum_{i,j}a_{ij}\phi_\lambda(i)(\text{log} a_{ij}^\prime - \text{log} a_{ij})
\end{equation}
$a_{ij}$ represent the probabilities in transition matrix of $\lambda$; $\phi_\lambda(i)$ is the probability of state $i$ in the stationary distribution of $\lambda$. (The stationary distribution will be further explained in section \ref{sec:learning-rep})
The \textit{Likelihood Distance} computes the distance based on both the state sequences and HMMs, where state sequence is a sample of the HMM (generative) model. The other method, \textit{Viterbi Distance}, computes the values by only comparing the HMMs. This makes Viterbi distance less susceptible to noises observed in state sequences, or in other words, less sensitive to small changes. This trade-off causes one method to perform better than the other depending on the targeted ground truth construct. 



\subsection{Learning Representations}
\label{sec:learning-rep}
We describe two methods for learning representations from the HMMs. The first method 
is interpretable and could be used for analyzing the data. The second method however gives better performance in predicting most of the constructs. 

\label{sec:stat}
 \textbf{Stationary Representation: } Each HMM is defined by a transition matrix. The transition matrix gives the probability of transitioning from one state to another, so $z_t T_{i}$ is the probability distribution for $z_{t + 1}$, and $\lim_{x\to\infty}z_t {T_{i}}^x$ is the probability distribution for $\lim_{t\to\infty}z_{t}$, which is the stationary distribution. Regardless of the starting state, the relative amount of time spent in each state is given by the stationary distribution, which is 
 the eigenvector corresponding to the largest eigenvalue of the transition matrix. We treat these stationary distributions as representations for time series, i.e., participants. 

\label{sec:spec}
\textbf{Spectral Representation: } A drawback of using stationary representation to represent participants is that it does not capture the relation between behavioral states, hence it might not be able to distinguish between participants with similar behaviors but that are ordered differently. In order to capture these differences we use the spectral representation, which is similar to spectral clustering \cite{von2007tutorial}. 
Specifically, we perform these steps: 
\textbf{1)} Calculate the distance matrix between participants using either the likelihood distance or the Viterbi distance described in section \ref{sec:clustering}
\textbf{2)} Compute the normalized Laplacian of the distance matrix 
\textbf{3)} Use $K$ largest eigenvectors (i.e., eigenvectors corresponding to largest eigenvalues) as representations of participants ($K$ is a hyperparameter). 
 \begin{table}[!t]
\centering
\begin{small}
\begin{tabular}{|c|l|}
\hline
\emph{Signal} & \multicolumn{1}{|c|}{\emph{Feature}} \\
\hline
\multirow{2}{10em}{\centering Biometrics: \\{\small 6 features} }
& Avg. Breathing Depth, Avg. Breathing Rate, Heart Rate  \\
& Std. Breathing Depth, Std. Breathing Rate, R-R Peak Coverage  \\
\hline
\multirow{4}{6em}{\centering Movement: \\{\small 15 features} }
& Intensity, Cadence, Steps, Sitting, Supine, Low G Coverage\\
& Avg. G Force, Std. G Force, Angle From Vertical \\
& Avg. X-Acceleration, Std. X-Acceleration, Avg. Y-Acceleration \\
& Std. Y-Acceleration, Avg. Z-Acceleration, Std. Z-Acceleration \\
\hline
\end{tabular}
\end{small}
\caption{Extracted Features from OMsignal.}
\label{feature}
\end{table}
\section{Data}
\label{sec:data}

Data used in this work comes from 
a study of workplace well-being that measures physical activity 
and physiological states of  hospital workers \cite{mundnich2020tiles}. 
The study recruited over 200 volunteers for ten weeks. 
Participants were 31.1\% male and 68.9\% female and ranged in age from 21 to 65 years. 
Participants held a variety of job titles: 54.3\% were registered nurses, 12\% were certified nursing assistants, and the rest with some other job title, such as respiratory therapist, technician, etc. 
Participants wore the sensors for different number of days, depending on the number of workdays during the study. Furthermore, participants exhibited varying compliance rates.  
Hence, the length of the collected data varies across participants. 
For this paper, we focused on 180 participants from whom at least 6 days of data was collected. 
In addition to wearing sensors, participants were also asked to complete surveys prior to the study. 
These pre-study surveys measured cognitive ability, personality and health states, which serve as ground truth constructs for our study. Constructs are shown in Table~\ref{tab:pred1-regress}. Data used in this paper was collected from a suite of wearable sensors produced by \textit{OMSignal Biometric Smartwear}. These OMSignal garments include sensors embedded in the fabric that measure physiological data in real-time and can relay this information to participant's smartphone. 
Table~\ref{feature} shows the sensor signals  
used in this work. 

\section{Results}
\label{sec:results}
We used  BP-AR-HMM with autoregressive lag $1$ to model the temporal data collected from sensors worn by the 180 high-compliance participants. 
For each participant, we 
used Z-score to normalize signals 
for data analysis, however since some statistical features such as mean and variance are useful for predicting constructs like age, both normalized and unnormalized signals were used in the prediction tasks in Table \ref{tab:pred1-regress}. 
The model, trained on normalized signals, identified 23 shared latent states describing  participants' behavior. Some of the states were only exhibited by a few participants. These rare states could convey useful information that helps identify noise  or anomalies in the data; however, their sparseness is not beneficial to the prediction and clustering tasks. Therefore, we ignore states observed in fewer than $5\%$ of the participants.

\subsection{Clustering}
\label{res:clus}
\begin{figure*}[t]
\centering
\includegraphics[width=\textwidth]{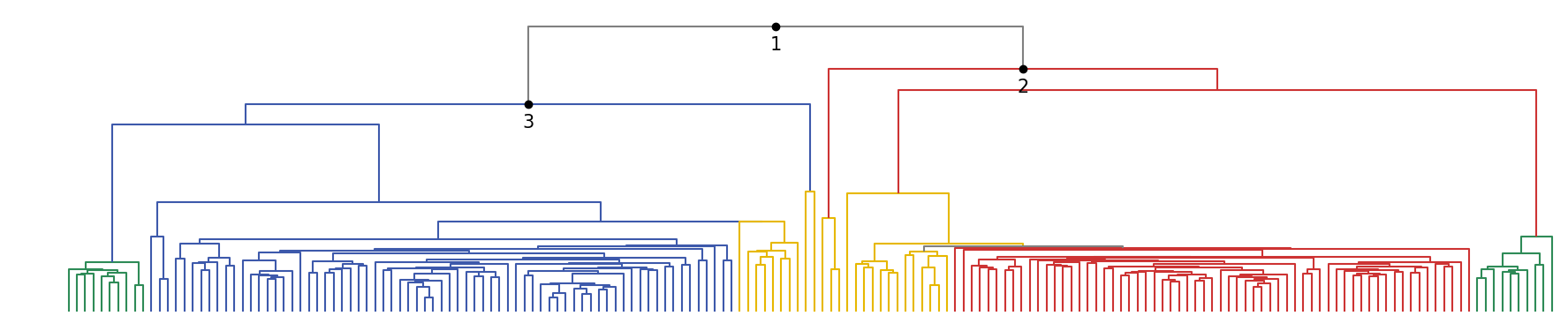}
\caption{Dendrogram showing the similarity of participants based on their learned states. }
\label{fig:clustering}
\end{figure*}
For validating the 
states learned by the model, we apply hierarchical agglomerative clustering on the distance matrix generated with likelihood distance. The resulting dendrogram is shown in Figure \ref{fig:clustering}. 
We used 
pre-study surveys to perform statistical tests and evaluate differences between the clusters.
We partitioned the dendrogram into clusters with more than five members by cutting the dendrogram horizontally on different depths. 
Based on the P-values obtained, 
the most important features differentiating the clusters were job type, age and gender in that order. This was aligned with our expectations, since different job types require different activities, also age and gender affect physiological signals. 

The first cut point (marked 1 in figure~\ref{fig:clustering}) separates registered nurses from other jobs types. 
The main difference between the two clusters (red and blue) is the frequency of three latent states, which we call \textbf{A}, \textbf{B}, and \textbf{C}. 
Variables related to acceleration and movement are almost zero for state \textbf{A}, which has a higher frequency for participants in cluster 3--non-nurse(blue). State \textbf{B} is more representative of higher activity levels and is more frequent for participants in cluster 2--nurses(red).   
 This also is aligned with our expectations, since the nurse occupation requires more activity compared to other job types in the study. 
State \textbf{C} mostly captures flexibility of work hours for non-nurses (non-nurse participants are more likely to finish their shifts earlier and have less than 12 hours worth of data in one shift). 
This clustering also 
separates participants based on their work shifts, day or night shifts. 
Work shifts are distinguished by state \textbf{D}. 
In this state, binary supine signal, which is activated when the participants is lying down, is on. 
It appears that state \textbf{D} captures quick naps in the workplace and has a higher frequency for night shift participants. Participants who exhibit state \textbf{D} are shown by color yellow in Figure~\ref{fig:clustering}. 

\subsection{Prediction}
We use the learned representations for each participant 
as features to predict the ground truth constructs. The objective is two-fold: not only do we want to predict, but also to gain understanding about what the latent states represent.
A possible way to understand latent behaviors is to quantify their importance in explaining constructs. The stationary representation described in section \ref{sec:learning-rep} has a clear interpretation, with each dimension representing the percentage of time spent in the corresponding state. 
We explain the behavioral states using the stationary representation with the following process: 
\textbf{1)} Get the stationary representations of participants
\textbf{2)} Run classification/regression on the representations to predict construct.
\textbf{3)} Retrieve the learned coefficients. 
\textbf{4)} Select the states with highest absolute coefficients and interpret these states based on their relation with the targeted construct.
Based on this approach, we recognized state \textbf{D}, described in the section \ref{res:clus}, as the most relevant state for differentiating between day and night shift employees. 
State \textbf{D} also has a high positive coefficient in predicting POS-AF and Well-being, whereas for hindrance stress it has a high negative coefficient. Hindrance stress is generally perceived as a type of stress that prevents progress toward personal accomplishments. Thus a plausible interpretation of this result is: Quick breaks during work hours could increase positive affect and well-being and decrease hindrance stress.


\begin{table*}[!t]
 \begin{adjustwidth}{-0.85in}{-0.85in}  
	\centering
    \scalebox{0.85}{
	\begin{tabular}{|c|c|c|c|||c|c||c|c||c|c||c|c|}
	\hline \hline
	 & 
	 & $ \rho $ & RMSE 
	 & $ \rho $ & RMSE 
	 & $ \rho $ & RMSE 
	 & $ \rho $ & RMSE 
	 & $ \rho $  & RMSE\\ \hline
\emph{Construct} & \textit{Description/Instrument}  &\multicolumn{2}{|c|||}{\emph{HMM-S}} & \multicolumn{2}{|c||}{\emph{HMM-SL}}&\multicolumn{2}{|c||}{\emph{HMM-SV}} &\multicolumn{2}{|c||}{\small{\emph{RWS}}} & \multicolumn{2}{|c|}{\small{\emph{Parafac2}}} \\ \hline
NEU    &  Personality: Neuroticism\cite{gosling2003very}                                        & 0.066  & 0.726     & 0.159 & 0.718     & \textbf{0.174} & \textbf{0.722}     & 0.048  & 0.728     & 0.116  & 0.724     \\\hline
CON   & Personality: Conscient.\cite{gosling2003very}                                         & -0.165 & 0.62      & \textbf{0.245} & \textbf{0.591}     & 0.181 & 0.6       & -0.033 & 0.613     & 0.093  & 0.612     \\\hline
EXT &   Personality: Extraversion \cite{gosling2003very}                                & 0.154  & 0.655     & 0.152 & 0.659     & \textbf{0.264} & \textbf{0.642}     & 0.178  & 0.65      & 0.038  & 0.66      \\\hline
AGR  & Personality: Agreeableness\cite{gosling2003very} & -0.428 & 0.491     & 0.122 & 0.485     & \textbf{0.191} & \textbf{0.479}     & 0.079  & 0.488     & 0.099  & 0.488     \\\hline
OPE & Personality: Openness \cite{gosling2003very} & 0.224  & 0.586     & 0.217 & 0.581     & \textbf{0.28}  & \textbf{0.571}     & 0.216  & 0.585     & -0.386 & 0.598     \\\hline
POS-AF  & Positive affect \cite{watson1988development} & 0.37   & 6.547     & \textbf{0.254} & \textbf{6.614}     & 0.231 & 6.686     & 0.139  & 6.821     & 0.112  & 6.822     \\\hline
NEG-AF  & Negative affect \cite{watson1988development}  & -0.278 & 5.293     & \textbf{0.235} & \textbf{5.139}     & 0.206 & 5.195     & 0.045  & 5.286     & 0.139  & 5.238     \\\hline
STAI &   Anxiety \cite{b23} & 0.016  & 8.975     & \textbf{0.196} & \textbf{8.817}     & 0.112 & 8.919     & 0.128  & 8.912     & 0.095  & 8.966     \\\hline
AUDIT   & Alcohol Use Disorders Test \cite{b24} & 0.1    & 2.159     & \textbf{0.362} & \textbf{2.017}     & 0.153 & 2.142     & 0.053  & 2.169     & 0.244  & 2.113     \\\hline
IPAQ & Physical activity\cite{b26}                                    & -0.57  & 15352 & 0.094 & 15191 & \textbf{0.115} & 15316 & 0.033  & 15311 & 0.097  & 15246 \\\hline
PSQI & Sleep quality \cite{buysse1989pittsburgh} & -0.682 & 2.366     & 0.178 & 2.318     & 0.142 & 2.33      & 0.193  & 2.322     & \textbf{0.194}  & \textbf{2.311}     \\\hline
Age & -  & \textbf{0.461}  & \textbf{8.613}     & 0.091 & 9.662     & 0.084 & 9.667     & 0.243  & 9.406     & 0.363  & 9.035     \\\hline
Health Limit  & Limitations due to health 
\cite{ware1992mos} & -0.75  & 23.284    & 0.196 & 22.704    & \textbf{0.333} & \textbf{21.986}    & 0.222  & 23.325    & 0.118  & 23.264    \\\hline
Emotional Limit & Limitations due to emotion \cite{ware1992mos} & -0.704 & 22.71     & \textbf{0.211} & \textbf{22.102}    & 0.164 & 22.504    & 0.042  & 22.652    & 0.091  & 22.553    \\\hline
Well being   & Psychological well-being\cite{ware1992mos}                                  & 0.077  & 18.458    & 0.152 & 18.302    & \textbf{0.276} & \textbf{17.904}    & 0.011  & 18.682    & 0.167  & 18.277    \\\hline
\small{Social Functioning}  & Social interaction ability\cite{ware1992mos} & 0.057  & 21.94     & 0.109 & 21.684    & 0.191 & 21.547    & 0.085  & 21.857    & \textbf{0.218}  & \textbf{21.541}    \\\hline
Pain & Index of physical pain\cite{ware1992mos} & 0.167  & 18.613    & 0.134 & 18.448    & \textbf{0.239} & \textbf{18.164}    & 0.023  & 18.658    & 0.102  & 18.571    \\\hline
\small{General Health} & Index of general health\cite{ware1992mos} & 0.211  & 17.062    & \textbf{0.27}  & \textbf{16.792}    & 0.171 & 17.28     & 0.151  & 17.311    & 0.2    & 17.105    \\\hline
\small{Life Satisfaction} &   Global life satisfaction\cite{diener1985satisfaction}    & -0.655 & 1.354     & 0.106 & 1.338     & \textbf{0.22}  & \textbf{1.317}     & -0.125 & 1.362     & 0.207  & 1.317     \\\hline
\small{Perceived Stress} & Perceived stress indicator\cite{cohen1994perceived} & 0.196  & 0.511     & 0.201 & 0.51      & \textbf{0.209} & \textbf{0.511}     & 0.195  & 0.513     & -0.728 & 0.524     \\\hline
PSY flexibility   & Ability to adapt
\cite{unknown}                              & -0.793 & 0.821     & 0.187 & 0.806     & \textbf{0.233} & \textbf{0.795}     & -0.077 & 0.823     & 0.103  & 0.813     \\\hline
PSY inflexibility   &  Inability to adapt
\cite{unknown} & -0.66  & 0.803     & \textbf{0.182} & \textbf{0.785}     & 0.152 & 0.79      & -0.013 & 0.803     & 0.006  & 0.8       \\\hline
WAAQ   &  Work Acceptance 
\cite{bond2013work} 
& \textbf{0.31}   & \textbf{5.65}      & 0.284 & 5.705     & 0.205 & 5.833     & 0.153  & 5.878     & 0.163  & 5.866     \\\hline
\small{Psych-Capital} &  
Psychological capital\cite{luthans2007positive} & \textbf{0.188}  & \textbf{0.656}     & 0.129 & 0.661     & 0.17  & 0.662     & 0.12   & 0.662     & 0.08   & 0.664     \\ \hline
\small{Challenge Stress} & Challenge stress indicator\cite{rodell2009can} & -0.639 & 0.622     & \textbf{0.171} & \textbf{0.615}     & 0.078 & 0.62      & -0.097 & 0.623     & -0.789 & 0.621     \\ \hline
\small{Hindrance Stress} &  Hindrance stress indicator\cite{rodell2009can} & 0.132  & 0.644     & 0.005 & 0.646     & \textbf{0.206} & \textbf{0.633}     & 0.035  & 0.647     & 0.143  & 0.637  \\   \hline \hline
	\end{tabular}
	}
	\caption{Evaluation of the model.
    The best performing model's results are highlighted in bold.}
\label{tab:pred1-regress}
\end{adjustwidth}
\end{table*}


\textbf{Quantitative Results: }For predicting constructs, 
we obtained stationary representations and spectral representations using both distance measures (likelihood and Viterbi distance). 
Spectral representation requires a hyperparameter K, the number of eigenvectors to include in the representation. We set the K to 10, 20, $\ldots$, 100 and use it to train the regression model.  
Since there are many ground truth constructs, one model can not be chosen for all the regression tasks. We use the ridge, kernel ridge and random forest regression on proposed representations and baselines and report the best model. 
The results are reported in correlation to the target construct ($\rho$) and Root Mean Squared Error (RMSE) using Leave-one-out cross validation. 
We compare our results against \textit{Random Warping Series (RWS)}\cite{wu2018random} and \textit{Parafac2}~\cite{harshman1970foundations}. RWS generates time series embeddings by 
measuring the similarity between a number of randomly generated sequences and the original sequence. 
This method has three hyper-parameters. 
Based on authors suggestion, we fixed the dimension of the embedding space to 512, 
and experimented with different values for the other two parameters. 
The second baseline we use is  \textit{Parafac2}~\cite{harshman1970foundations}. This approach views the data as a tensor (3 dimensional array) of participants-sensors-time and decomposes it into hidden components.
For Parafac2 the number of hidden components is a hyper-parameter. 
We varied the number of hidden components from one to ten and report the best results. 
Overall, the results of the predictions based on the HMM's latent states were systematically better, outperforming the baseline method in 25 of the 27 constructs predicted. It's worth mentioning that, except for HMM-S, which is non-parametric, all other four models in Table \ref{tab:pred1-regress} have hyperparameters that need to be set. We tune the hyperparameters by running 10 different settings and selecting the setting with best results. 
Between our own representations HMM-Stationary, HMM-Spectral-Likelihood and HMM-Spectral-Viterbi (HMM-S, HMM-SL and HMM-SV respectively), HMM-SV performs better for some construct while HMM-SL gives better results for others. This could be because of differences between Viterbi and likelihood distance's sensitivity to small variations in the data (discussed in section \ref{sec:clustering}). Also, HMM-S is not a good representation for prediction and is better suited for analysis of the data.

\section{Conclusion}
We described a method for learning behavioral representations from physiological data captured by wearable sensors. 
We used this framework to model 
data collected from 
workers in a 
hospital. 
The latent states learned by the model 
can be used to predict their self-reported health and psychological well-being. In comparison to alternative models, our framework improves performance with compact representations of the multivariate time-series, leading to less overfitting and easier interpretation of the states learned.  
Concluding, we show that this framework can also cluster study participants into meaningful groups. 
This work can be extended in a number of ways. 
One possible direction is making this framework supervised. Using our current framework helps in analyzing data, but 
making this framework supervised could be more suited for a prediction task. 

\textbf{Acknowledgements}
The research was supported by the Office of the Director of National
Intelligence (ODNI), Intelligence Advanced Research Projects Activity
(IARPA), via IARPA Contract No 2017-17042800005.

%
%
\bibliographystyle{splncs04}
 
\bibliography{mybibliography}
 
%




\end{document}